# Smart Spatial Planning in Egypt: An Algorithm-Driven Approach to Public Service Evaluation in Qena City

\* Mohamed Shamroukh Mohamed, \*\* Mohamed Alkhuzamy Aziz
(\*) Faculty of Arts, South Valley University, Egypt (\*\*) Faculty of Arts, Fayoum University, Egypt
Correspondent: Mohamed Shamroukh Mohamed, South Valley University, Egypt,
E-mail: mohamedshamroukh@art.svu.edu.eg, ORCID ID: https://orcid.org/0000-0002-1005-2600

**Abstract**:

National planning standards for public services in Egypt often fail to align with unique local characteristics. Addressing this gap, this study develops a tailored planning model for Qena City. Using a hybrid methodology—descriptive, analytical, and experimental—the research utilizes Python programming to generate an intelligent spatial analysis algorithm based on Voronoi Diagrams. This approach creates city-specific planning criteria and evaluates the current coverage of public facilities. The primary contribution of this study is the successful derivation of a localized planning standards model and the deployment of an automated algorithm to assess service efficiency. Application of this model reveals a general service coverage average of 81.3%. Ambulance stations demonstrated the highest efficiency (99.8%) due to recent upgrades, while parks and open spaces recorded the lowest coverage (10%) caused by limited land availability. Spatial analysis indicates a high service density in midtown (>45 services/km²), which diminishes significantly towards the outskirts (<5 services/km²). Consequently, the Hajer Qena district contains the highest volume of unserved areas, while the First District (Qesm 1) exhibits the highest level of service coverage. This model offers a replicable framework for data-driven urban planning in Egyptian cities.

## 1. Introduction:

Public services [1] are an essential part of a city's structure. The availability and sophistication degree of such services are fair measures of progress for any city. In this context, Geographic information systems "GIS" offers solutions that support the decision-making processes regarding management, planning and distribution of services, ultimately improving the standard of living in cities (Aziz, 2007, p. 11). Investigating services planning standards is one of the most relevant issues concerning human progress regarding its proper definition and needs. Planning standards can be reconsidered by studying the variation in the distribution of geographical phenomena and the characteristics of geographic areas. More effort should be exerted in defining these standards parallel to the characteristics of each region. Such efforts will facilitate appropriate allocations of services and accurate definitions of future developmental efforts.

The problem of the study is that the planning standards are not suitable for the characteristics of the Egyptian cities, which include more population and intensive daily use of services. The solution to this problem is to create new planning standards that suit the rapidly changing nature of cities, and to generate these criteria current services and their intensity and the built-up areas are going to be used to reflect the characteristics of the city, taking this abroach is a new way to generate such criteria.

This study attempts to derive planning standards for public services in the city of Qena that are compatible with the characteristics of the city, the geographical distribution of the population, the built-up area, and the services therein. This could be accomplished by dividing the city into regions according to the geographical distribution of services using Voronoi Analysis [2] and each part creates a polygonal shape representing the area of attraction for the service from which the coverage distance [3] for each service is measured.

---

(1) public services includes (educational, health, religious, cultural, and recreational, postal, fire stations).

[2] The origin of the name Voronoi diagram goes back to the Ukrainian mathematician Georgi Fedosevic Voronoi in 1905 AD. However, this type of diagram was used before the name "Voronoi" was coined in 1644 AD by Rene Descartes. It was also used in 1850 AD by the German mathematician Johann when he applied it to represent binary data Dimensions and 3D (Kang, 2016, p. 1233).

[3] The coverage distance is the maximum distance for each service to cover, and it can be derived from Voronoi polygons or directly from the national planning criteria, the study suggests that the average coverage distance from the Voronoi analysis of each public service is used as the maximum service limit of the proposed planning standards.

Then, the average coverage distances are calculated for it, from which the most appropriate standard for the service is determined. Accordingly, the extracted standard expresses the service within its geographical scope considering the areas of population intensity, services, and the least concentrated areas. After that an algorithm is built to apply these standards to the city to determine the coverage percentage for each service and administrative section.

What are the most compatible criteria for public services in Qena city? What is the best way to develop new standards for the city?

## 2. Data:

### 2.1. Study Area:

The city of Qena lies in the middle of markaz Qena and Qena Governorate, as shown in Figure (1). It is bordered on the South by the Nile River and on the North by the new city of Qena. While on the East, it is bordered by the village of Al-Saleheya, and the new city of Qena. Qena is bordered from the West by the village of Dandara and the desert of Qena. Moreover, the city of Qena is located between Longitudes 32° 40' 50", 32° 47' 33" E and between latitudes 26° 7' 57", 26° 13' 43"N.

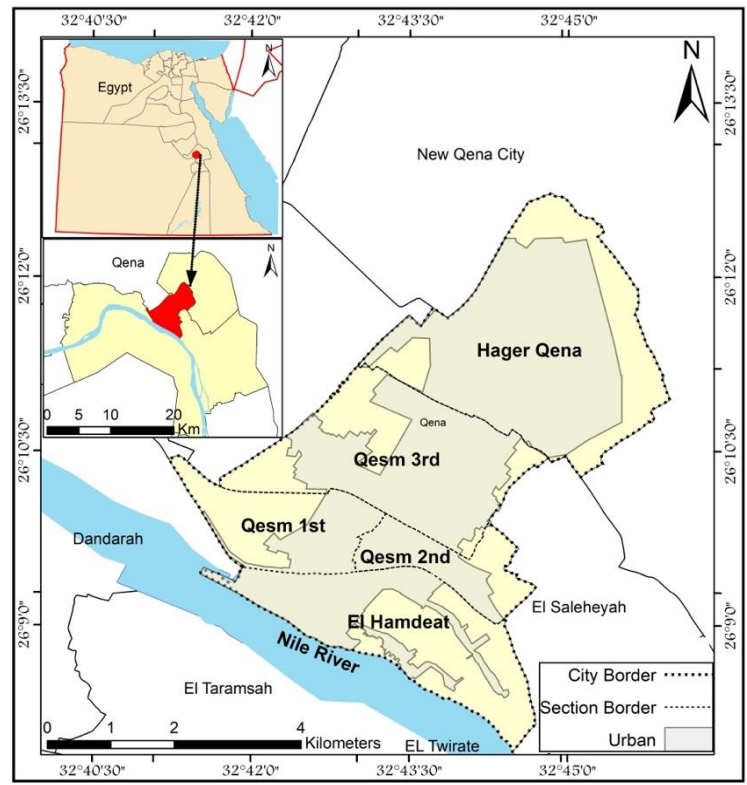

Figure (1): study area
Source: Created by Authors

The area of Qena city is about: 29.85 km², representing 11% of the area of markaz Qena. The city's built-up area represents about 19 km² or 63.5% of the area. Furthermore, the city of Qena consists of five sections: Hajer Qena, which represents 11.12 km² of the total area, followed by the Qesm third representing an area of 7.3 km², followed by Al-Hamdeat with an area of 6.1 km², followed by the Qesm1st with an area of 3.22 km² and finally is Qesm2nd with an area of 2.11 km².[4]

The population of Qena city is about: 238056[5] representing 7975/km2. Furthermore, the city consists of five sections: Hajer Qena, with a population of 9345, Qesm third with a population of 111715, and Al-Hamdeat with a population of 47335, Qesm1st with a population of 36164 and finally is Qesm2nd with a population of 33496.

## 3. Methodology:

### 3.1. *Voronoi Smart Spatial Analysis*

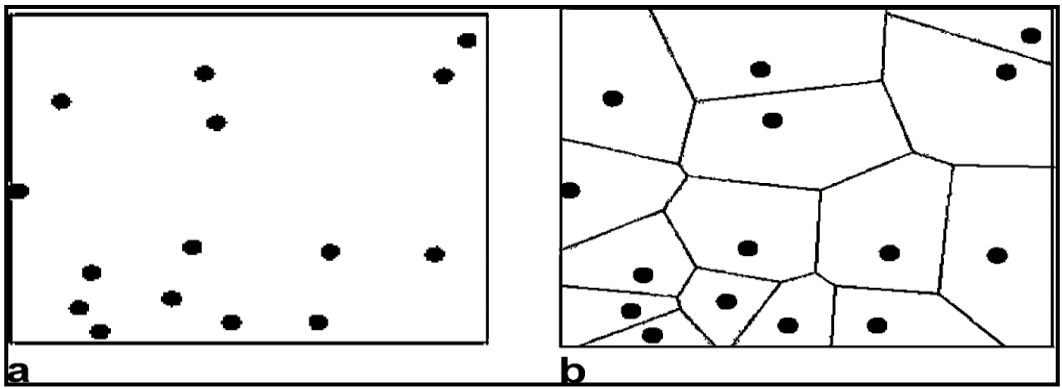

Source: Encyclopedia of GIS.Voronoi Diagram,2008, p.1233.

Figure (2): Voronoi Diagram

Figure (2) shows the Voronoi diagram analysis that includes a set of features within an area to form a set of shapes(polygons); the polygon represents a shape surrounding the phenomenon where any point within the polygon of the phenomenon is closer to it than any other phenomenon (Kang, 2008, p. 1232). The Voronoi diagram can be defined as dividing the place into sectors or ranges depending on the distance between certain features of a particular phenomenon. These are Voronoi points, whose surrounding spaces are consequently areas of influence. Thus, any spatial spot within the phenomenon catchment area[6] is closer to its Voronoi point than any other Voronoi point.

---

[4] The areas were calculated using ARC/GIS 10.7.1
[5] Information and Decision Support Center in Qena Governorate2021.

Applying the Voronoi Analysis to the public services in Qena city requires transforming the point features "service" into a set of polygons. Each polygon represents the Catchment Area of the service, or the measurable coverage range, whose geographical limits determine residents' eligibility to get services. It varies according to the area of the polygon. Thus, its more efficient to use average coverage distances for the service in the city.

The algorithm [7] built using the Python programming language and the Arcpy Library developed by (ESRI). Applying the Voronoi analysis Algorithm saves both time and effort, as it is usually performed over ten services in about 12 seconds, with 20 output files. Once the folder containing the services is selected, Analysis is done automatically. Repeating the analytical processes automatically eliminates the need to specify the inputs and outputs for each file separately.

### *3.2.    Intelligent spatial Analysis Algorithm for Public Services*

An algorithm [8] is a clear, precise, unambiguous, mechanically executable sequence of basic instructions, usually intended to achieve a specific purpose (Erickson, 2019, p. 1). In other words, an algorithm is a procedure to accomplish a specific task and is the idea behind any computer program. It must solve a specific problem (Richard Szeliski, 2020, p. 3). An algorithm is also a set of rules that leads to solving a complex problem (Kennedy, 2000, p. 2).

Geospatial data analysis is a critical component of decision-making and planning for numerous applications. Geographic Information Systems (GIS), such as ArcGIS ® and QGIS [9], provide rich analysis and mapping platforms. Modern technology enables us to collect and store massive amounts of geospatial data. However, the data formats vary widely, and Analysis requires numerous iterations. These characteristics make computer programming essential for exploring this data. Python is an approachable programming language for automating geospatial data analysis. (Tateosian, 2015, p. 1). Applying intelligent spatial analyses based on the Python language saves time and effort and automates the workflow.

---

[6] A catchment area is the area which surrounds the service, and it can be specified by the local criteria or the international criteria.
[7] Source code GitHub repository: https://github.com/MohamedShamroukh/VoronoiAnalysis
[8] Al-Khwarizmi is a word derived from the Latin word algorithms, which in turn derives from the name Al-Khwarizmi, a Muslim mathematician, astronomer, and geographer. He is the founder of algebra and has many contributions to geography.
[9] QGIS a free, open-source geodata processing package that works on various operating systems.

Figure (3) shows the design scheme of the data flow and processing operations in the intelligent spatial analysis algorithm for public services according to the proposed planning standards of the city. Moreover, the analyses are applied automatically, as mentioned above. The proposed planning standards are thus saved within the Script, enabling the algorithm to determine the standard corresponding to the service in question. The analytical process proceeds without needing to refer to the user in each step. Furthermore, the algorithm includes several steps to determine the served and unserved areas, as indicated below:

(1) Determining the corresponding standard for each service.
(2) Applying spatial coverage analyses corresponding with defined city borders.
(3) Defining enforcement of minimum / maximum planning standards from the city's built-up area.
(4) Determining served and less-served areas.
(5) Point out areas for each case, filing it.
(6) saving them in different graph-format jpg files.
(7) Defining service geographical distribution patterns.

All the steps mentioned above takes only around 5 seconds. The intelligent algorithm can, therefore, identify and make decisions about processing operations with minimal time and effort, by applying overlapping analysis to the built-up area layer and the service coverage layer, it was possible to derive the serviced areas from the built-up areas.

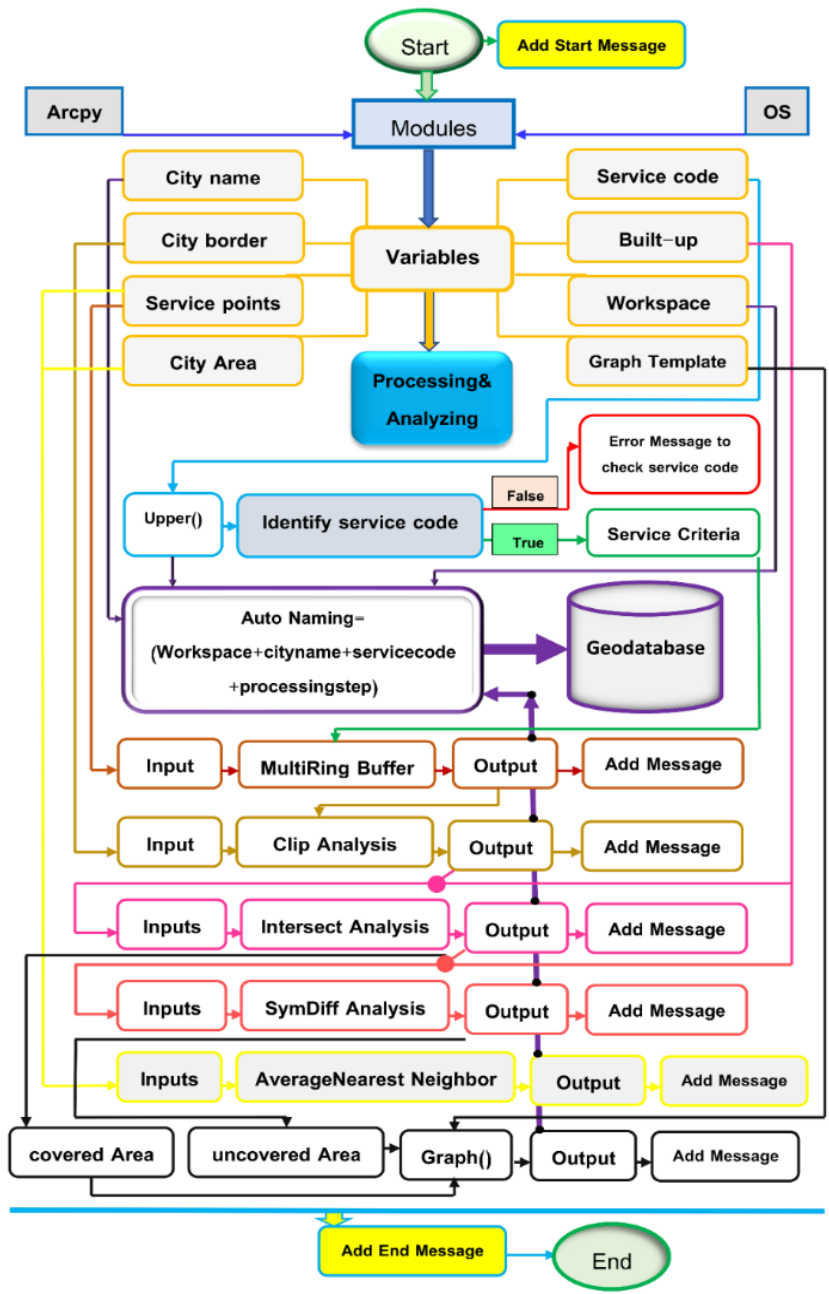

Figure (3): The design diagram of the Smart Spatial Analysis Algorithm

Source: Created by Authors

the source code [10] of the Smart Spatial Analysis Algorithm for public services according to the proposed planning standards for the city of Qena. In addition to the graphical user interface (GUI) of the algorithm in Figure (4), termed after its analytical function and source of planning standards (i.e., Qena Smart Spatial Analysis for Services). Furthermore, the algorithm can be applied and implemented within the ARC/GIS-Pro program environment through the (GUI), or it could be applied independently within an IDE. The application process requires defining a set of variables through which the inputs and outputs are determined, namely:

- Variable City Name: It is an optional variable that adds the city's name to the output of the Analysis to distinguish between cities.
- Variable Service code specifies the service to which the analyses will apply. User inputs 0are unified so that the algorithm can identify the service, determine the corresponding standard, and then apply the analyses.
- Variable Input point service layer: the user defines the point layer of the service to which the analyses will be applied.
- Variable input city border layer: This defines the city border layer to be used later as a constraint for the output range.
- Variable input city built-up layer defines the urban block layer of the city, which determines the service coverage of the city's built-up area.
- Variable: input City Area: specifies and determines the area of the city; To accurately apply statistical analyses.
- Variable: Workspace (Geodatabase): It defines the Geodatabase that will save all the algorithm's outputs.
- Variable: input Graph Template: which determines the graph template.

---

[10] Source code GitHub repository: https://github.com/MohamedShamroukh/SmartSpatialAnalysis

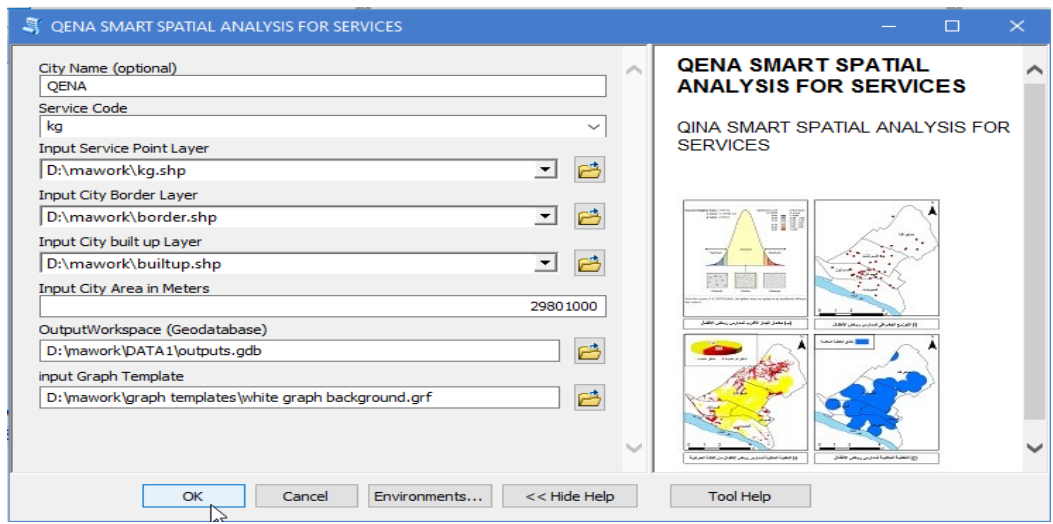

Figure (4): The graphical user interface of the intelligent spatial analysis algorithm for public services

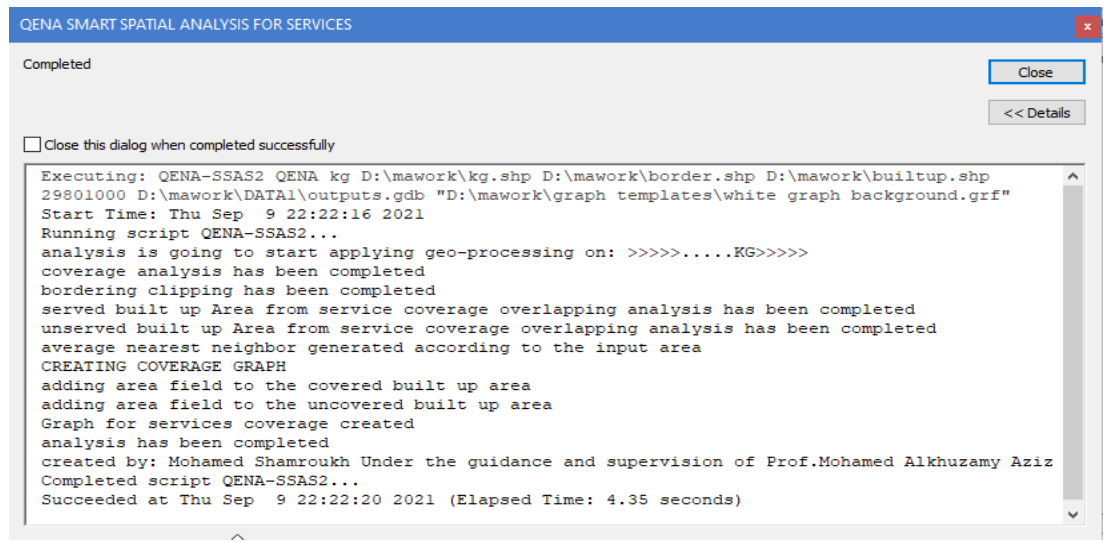

Figure (5): Implementation report of the Smart Spatial Analysis Algorithm for public services.

Figure (5) shows the implementation of the Algorithm Report for spatial Analysis of public services according to the proposed planning standards. This report shows the operational steps determined in the Algorithm Source Code. They appear as messages shown to the user while applying the analyses at the bottom of the text report. Such messages indicate the analyses' type and duration in fine prints in the margins of the textual report (typically 4.350 seconds). The following are some prominent features of the algorithm:

- The ability to automate the processing of Big Geodata [11]
- Rapidity in implementing processing operations and spatial Analyses l (typically just a few seconds).
- Automatic labelling of outputs, based on types of inputs.
- Identifying services' corresponding standards from available internal data.
- Relying on the Python language and open-source software facilitates development.
- Ability to work independently "Stand Alone Algorithm".
- Ease of shareability as an Arc-Toolbox or Share-Your-Script.
- The ability to develop the algorithm to suit any region around the world.

---

[11] The concept of Big Geodata is comparatively recent and deals with the well-defined and essential subset of Big geodata. As the Landsat, volume is no stranger to geodata, and today our ability to collect and acquire geodata vastly exceeds our ability to store or process them. (Goodchild, 2017, p. 20).

## 4. Results:

### *4.1. planning standards creation*

The proposed planning standards model for public services in Qena can build-able through the following procedures:

- Voronoi Analysis: by determining the catchment areas and calculating the average of spatial coverage distances for each of the educational, health, religious, postal, and fire station services.
- Approving a regional or international standard: The local standard for cultural services represented in cultural palaces reaches 250,000 people per unit or slightly more. The regional standard represents spatial coverage of 3-5 km/unit. Local standards set a standard of 7 m$^2$/person of parks and open areas and seek to reach 10 m$^2$/person, while international standards set 10-12 m$^2$/person. Therefore, the average international standard can be proposed, i.e., 11 m$^2$/person for parks and open spaces.

#### *4.1.1. Educational services*

- Primary Schools:

Figure (6-a) shows a Voronoi analysis of primary schools in Qena. The catchment areas for primary schools vary, reaching the maximum area in the "Al-Seka Al-Hadid" school, whose catchment area is 2.7 km$^2$ and a coverage distance of 2.2 km. The lowest is in Al-Manshiya school, with an area of 0.49 km$^2$ and a coverage distance of 0.31 km. By calculating the average, it reached about 0.753 km.

- Kindergarten schools:

Figure (6-b) shows a Voronoi analysis of the kindergarten schools in Qena. The catchment areas for kindergarten schools vary, reaching the maximum area in "Al-Seka Al-Hadid" school, which has a catchment area of 2.7 km$^2$, and a coverage distance of 2.2 km. At the same time, the lowest area is in Al-Manshiya School, with an area of 0.49 km$^2$ and a coverage distance of 0.310 km. By calculating the average, it reached about 0.715 km.

- Secondary schools:

Figure (6-c) shows a Voronoi analysis of the secondary schools in Qena. The catchment areas for secondary schools vary, reaching their maximum in the Nile School, which has a catchment area reaching 6 km², and a coverage distance of 4.16 km. At the same time, the lowest catchment area for secondary schools is in Qena Al-Azhar Institute, with an area of 0.22 km² and a coverage distance of 0.998 km. By calculating the average, it reached about 1.097 km.

- Preparatory schools:

Figure (6-d) shows a Voronoi analysis of the preparatory schools in Qena. The catchment areas for the preparatory schools vary, reaching the maximum in the Safwa school, which has a catchment area of 3 km² and a coverage distance of 2.1 km. At the same time, the lowest area is in Sidi Abdel Rahim school, which reaches 0.064 km², with a coverage distance of 0.46 km. By calculating the average, it reached about 0.880 km.

It is clear from the results of Voronoi's Analysis of educational services in the city of Qena that the catchment areas are small in the downtown areas; this is due to the concentration of the services and the population in the midtown (center of the city). On the other hand, the areas of attractions for educational services increase towards the outskirts due to the decrease in the number of services and their connections with the geographical distribution of the population.

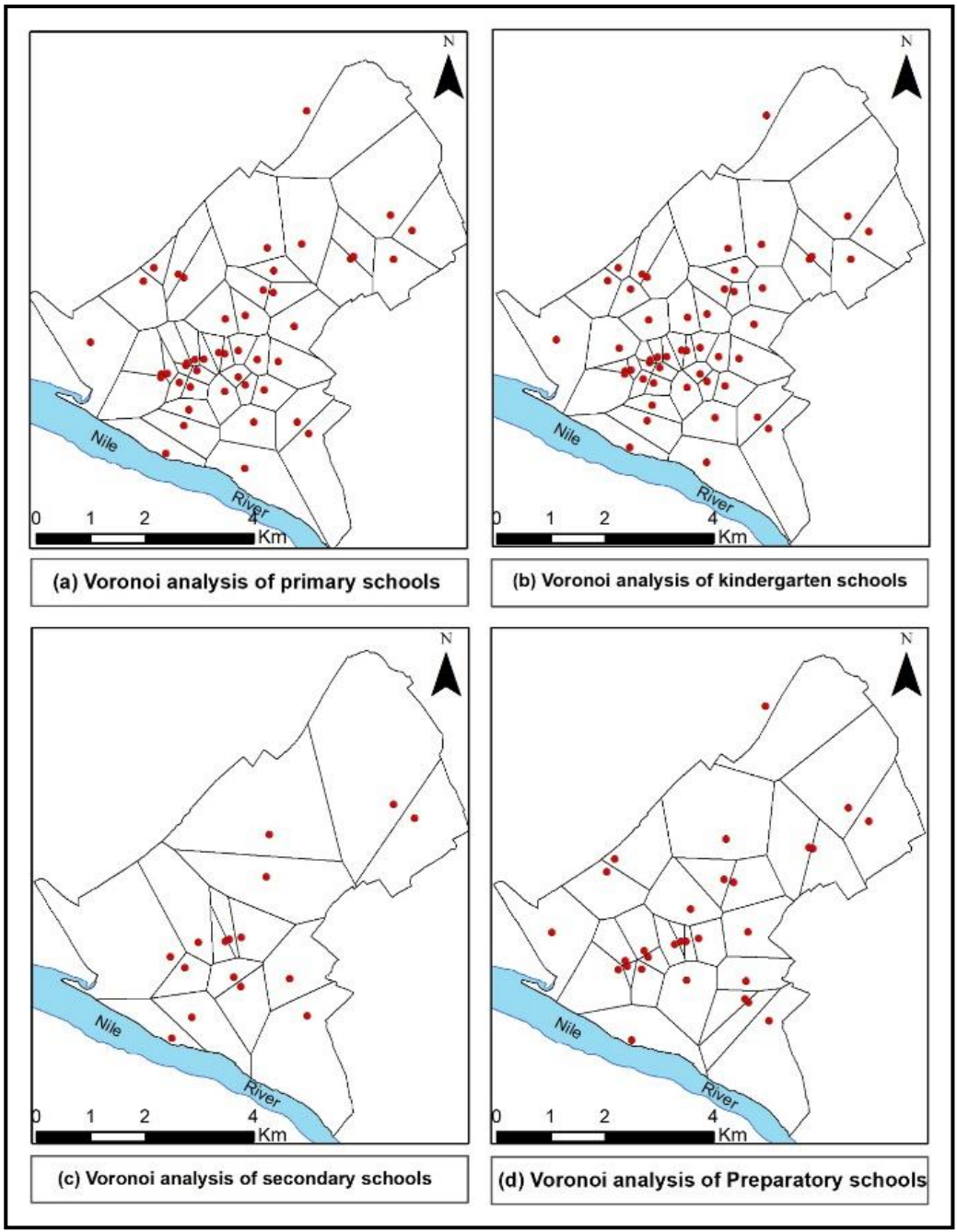

Figure (6): Voronoi Analysis of educational services in Qena city in 2021

*4.1.2. Health services*

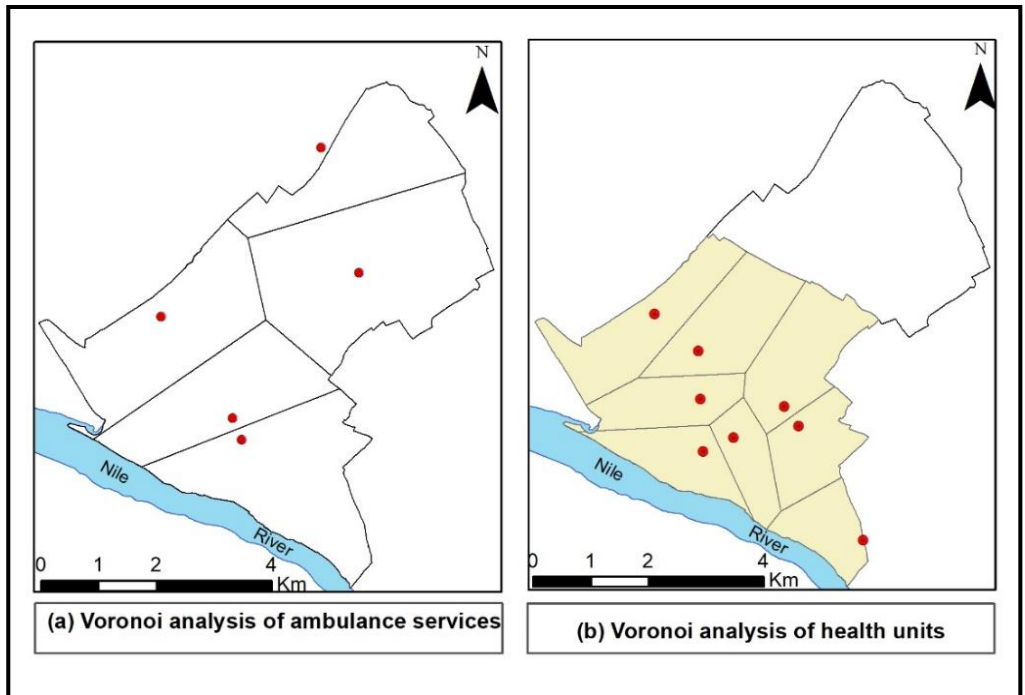

Figure (7): Voronoi Analysis of health services in Qena city in 2021

- Ambulance:

Figure (7-a) shows a Voronoi analysis of the ambulance points in the city of Qena. The catchment areas for the ambulance stations vary in their areas, reaching their maximum area at Al-Konouz ambulance station, with a catchment area of 8.6 km$^2$ and a coverage distance of 4.2 km, due to the existence of South Valley University, which has its ambulance station. On the other hand, the lowest catchment area is in the 6$^{th}$ km ambulance station, with an area of 4.4 km$^2$ and a coverage distance of 3.7 km. The Analysis shows a good distribution of the service. By calculating the average, it reached about 2.456 km.

- Hospitals and health units:

Figure (7-b) shows Voronoi's Analysis of health units in Qena in the administrative divisions of (Al-Hemadeat) and (first, second, and third divisions). Moreover, catchment areas for Health Units vary, reaching a maximum in Al-Maana's Health Unit, which has a catchment area reaching 3.4 km$^2$ and a coverage distance of 3.6 km. This is due to its location on the city's outskirts, where the population concentration is low. On the other hand, the lowest catchment area for hospitals and Health Units is in Qena General Hospital, with an area of 1.4 km$^2$ and a coverage distance of 2 km, due to the concentration of health services in the old midtown. By calculating the average, it is about 1.076 km. The Analysis was not applied to specialized and educational hospitals, as they are regional services.

### 4.1.3. Religious services

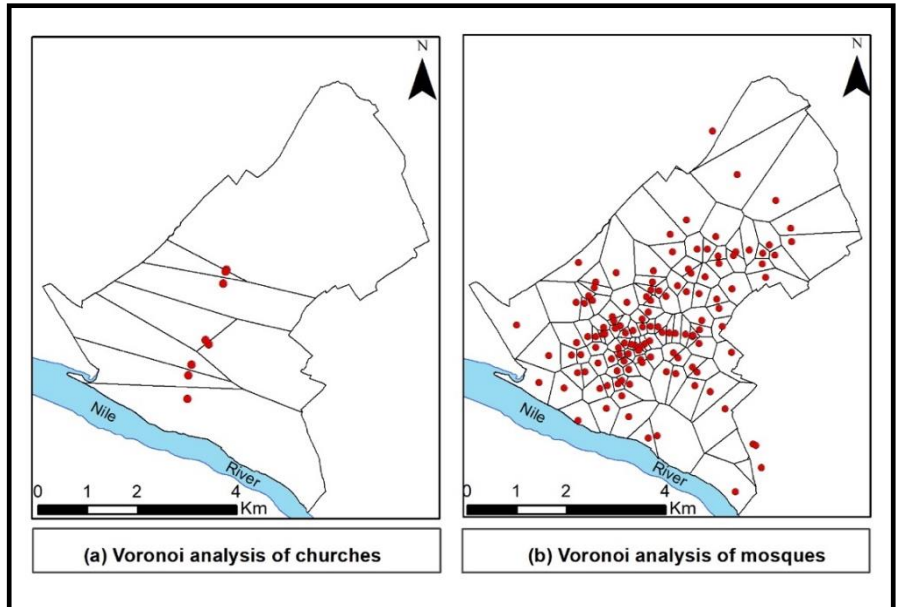

Figure (8): Voronoi Analysis of religious services in the city of Qena in 2021

- Churches:

Figure (8-a) shows a Voronoi analysis of the churches in Qena. The catchment areas for churches vary, reaching a maximum in the "Peter and Paul" church, which has a catchment area reaching 13.4 km$^2$, with a coverage distance of 4.6 km. The size of this church's catchment area includes the South Valley University area and the desert areas in the northern parts of the city. At the same time, the lowest area is in Mary Peter Church, with an area of 0.609 km$^2$ and a coverage distance of 1.9 km. By calculating the average, it reached about 2.032 km.

- Mosques:

Figure (8-b) shows a Voronoi analysis of mosques in the city of Qena. The area for mosques varies, with the maximum area being in the mosques on the outskirts, with an average catchment area of 2 km$^2$ and a coverage distance of 2 km. At the same time, the lowest area is in the midtown of Qena, with an area of 0.02 km$^2$ and a coverage distance of 0.185 km. It is clear from the Analysis that the service areas are small in the downtown areas, and the Analysis shows their expansion towards the outskirts. By calculating the average, it reached about 0.414 km.

*4.1.4. Other services*

- Fire Stations:

Figure (9-a) shows a Voronoi analysis of the fire stations in Qena city. The catchment areas for the fire stations vary, with the maximum being in the main Qena fire station area, which has a catchment area of 15 km² and a coverage distance of 4.9 km. On the other hand, the lowest is in a downtown fire station with an area of 14 km² and a coverage distance of 4.6 km, and by calculating the average, it reached about 2.9 km.

- Post offices:

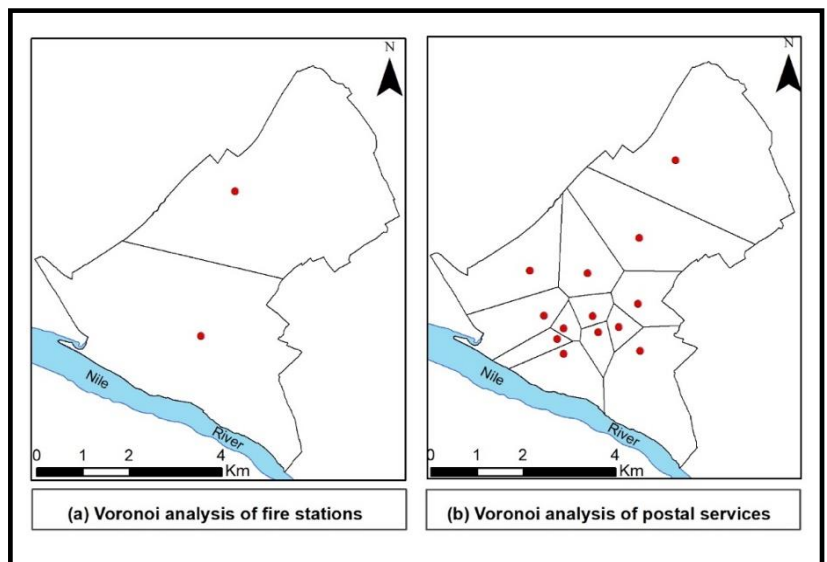

Figure (9): Voronoi Analysis of postal services and fire stations services in Qena in 2021

Figure (9-b) shows a Voronoi analysis of the post offices in Qena. The catchment areas vary; the maximum area is in South Valley University's post office, with an area of 8.8 km² and a coverage distance of 4 km. In comparison, the lowest area is near Qena Security Directorate's Square, which has a catchment area of 0.321 km² and a coverage distance of 0,702 km. The Analysis shows the concentration of postal services, the compactness of the catchment areas in the center of the city, and their expansion in the outskirts, due to its connection with the concentration of the population. By calculating the average, it reached about 1.5 km.

## 4.1.5. The Generated Planning Standards

Table (1) shows the proposed standards as the analysis result for the public services in Qena city using Voronoi model, which indicates both the minimum and maximum distances required for each public service as follows:

- Minimum Service Limit: represents the minimum distance of the service, which is necessary for this service to exist, if this distance is less than this size, its existence is uneconomic, the extent of the minimum distance differs from its counterparts of the same level when the population density increases and decreases from one location to another.
- Maximum Service Limit: represents the maximum distance of the service, it affects the efficiency of the service. Furthermore, it is necessary to provide another service of the same type and level, to match the increasing size of the population provided with such a service.

Table (1) The proposed planning standards for public services in Qena City

| No. | Group | Service | Min (km) | Max (km) |
|---|---|---|---|---|
| (1) | Educational services | Kindergarten | 0.358 | 0.715 |
| | | Primary schools | 0.377 | 0.753 |
| | | Preparatory schools | 0.440 | 0.880 |
| | | Secondary schools | 0.548 | 1.097 |
| (2) | Health services | Ambulance | 1.228 | 2.456 |
| | | Health units | 0.538 | 1.076 |
| | | Hospitals | 40-50 | |
| (3) | Religious services | Mosques | 0.207 | 0.414 |
| | | Churches | 1.016 | 2.032 |
| (4) | Cultural and recreational services | Libraries and cultural centres | 250 thousand people or more / 3-5 km | |
| | | parks and open areas | 11 m²/person | |
| (5) | Other services | Postal services | 0.752 | 1.504 |
| | | Fire extinguishing points | 1.440 | 2.880 |

Source: results of the Voronoi Analysis of Qena Services using ARC/GIS and PyCharm

## 4.2. Assessment of public services according to the proposed planning standards

### 4.2.1. Educational services

Education is the first sign of progress and development of societies, and it is a criterion for measuring their progress. Providing scientific, technical, and professional knowledge contributes to building society in all fields including economic, social, political, cultural, and technological that depends on education (Al-Dulaimi, 2015; p. 87), and education has always been an important part of the development of societies as a variable factor. Moreover, the development of various educational facilities contributes significantly to urban development (Nuzir & Dewancker, 2014, p. 632).

- Kindergarten:

Figure (10-a) shows the spatial coverage of kindergarten schools in the city from the built-up area according to the proposed standards. The serviced areas according to the standard represent about 87% of the built-up area, while the unserved areas represent about 13%, including South Valley University. Figure (10-b) shows the coverage areas for each administrative division; Qesm3$^{rd}$ is the first one with a coverage area of 97%, followed by Qesm2$^{nd}$ by 96%, then Qesm1$^{st}$ by 95%, then (Al-Hemadeat) by 85%, finally (Hajer Qena) with 68% of the built-up area. On the other hand, unserved areas reached a maximum of 32% in (Hajer Qena), And the lowest area is in Qesm3$^{rd}$ at 3% of the built-up area.

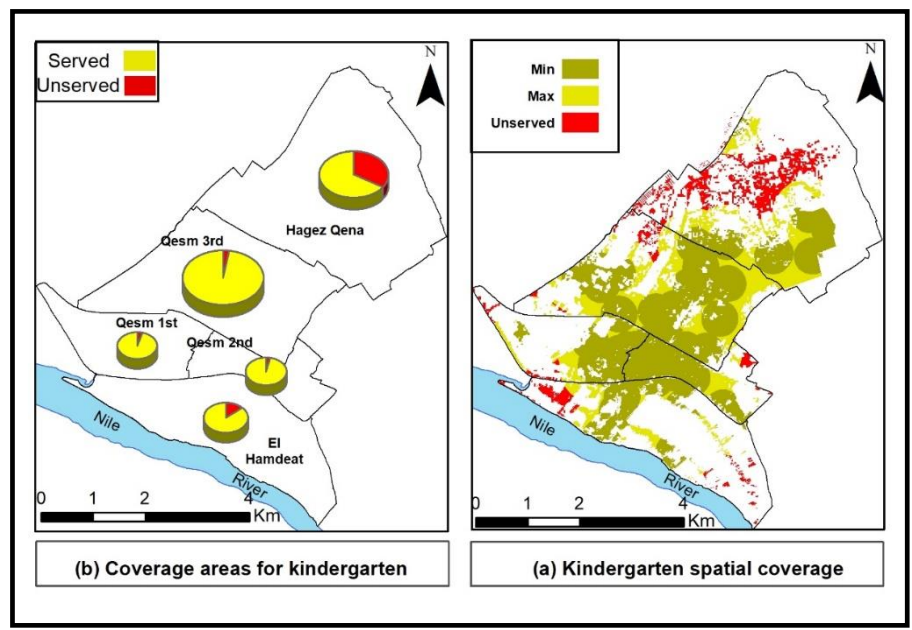

Figure (10): Assessment of kindergarten schools in Qena in 2021

- Primary Education:

Figure (11-a) shows the spatial coverage of primary schools in Qena from the city's built-up area according to the proposed standards. The serviced areas according to the standard represent 87% of the built-up area, while the unserved areas represent 13%, including the South Valley University. Figure (11-b) shows the coverage areas for each administrative division.

Qesm2nd is the first in the percentage of coverage for the serviced areas with 97 %, followed by Qesm3rd with 96%, then Qesm1st, 94%, and Al-Hemadat, 86%. Finally, Hajer Qena with 69% of the built-up area. There are non-serviced areas reached their maximum in Hajer Qena with 31% and their minimum in Qesm2nd with 3% built-up area.

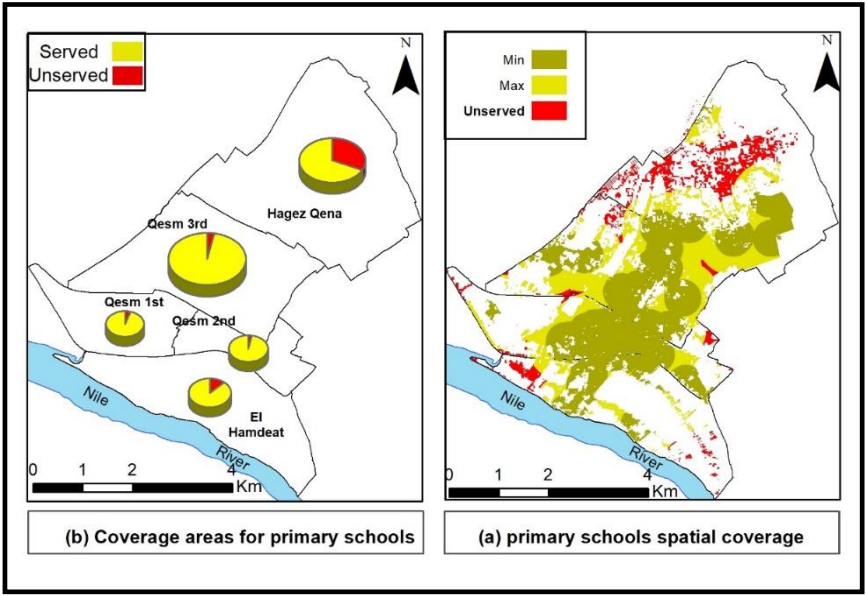

Figure (11): Assessment of primary schools in Qena in 2021

- Preparatory Education:

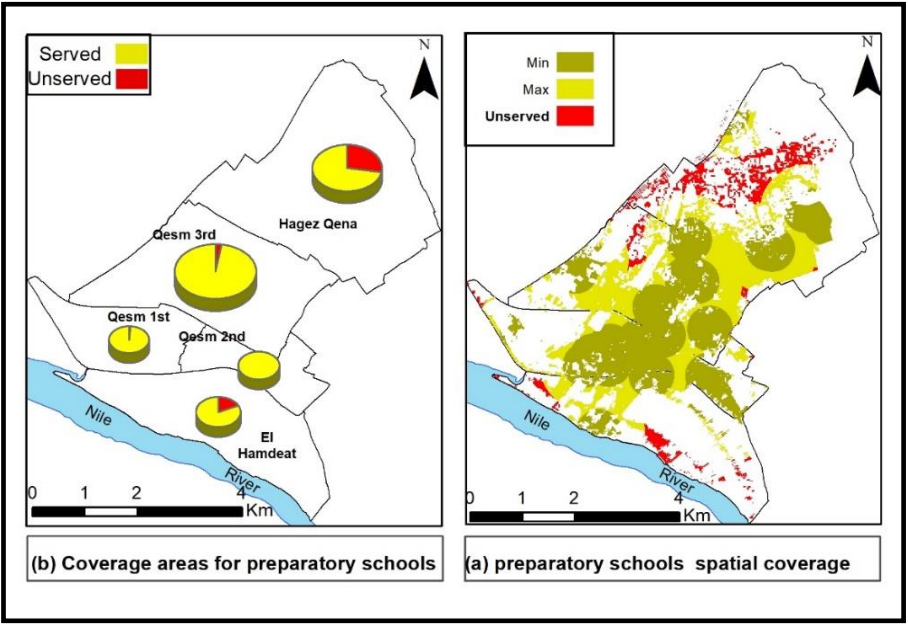

Figure (12): Assessment of preparatory schools in Qena in 2021

Figure (12-a) shows the spatial coverage of the preparatory schools in Qena from the built-up area according to the proposed standards. The serviced areas according to the standard represent 88.5% of the total urban mass of the city, while the unserved areas represent 11.5%, including the South Valley University. Figure (12-b) shows the coverage areas for each administrative division so that Qesm1st and Qesm2nd are in the first rank in the coverage area for preparatory schools with a coverage percentage of 99% of the serviced areas, followed by Qesm3rd with a coverage percentage of 97%, then (Al-Hemadat) with 81%, and finally (Hajar Qena) with 73% of the built-up area. On the other hand, the unserved coverage area for the preparatory schools is the highest in Hajer Qena (with 27%) and the lowest in Qesm1st and Qesm2nd (with 1% of the built-up area).

- Secondary Education:

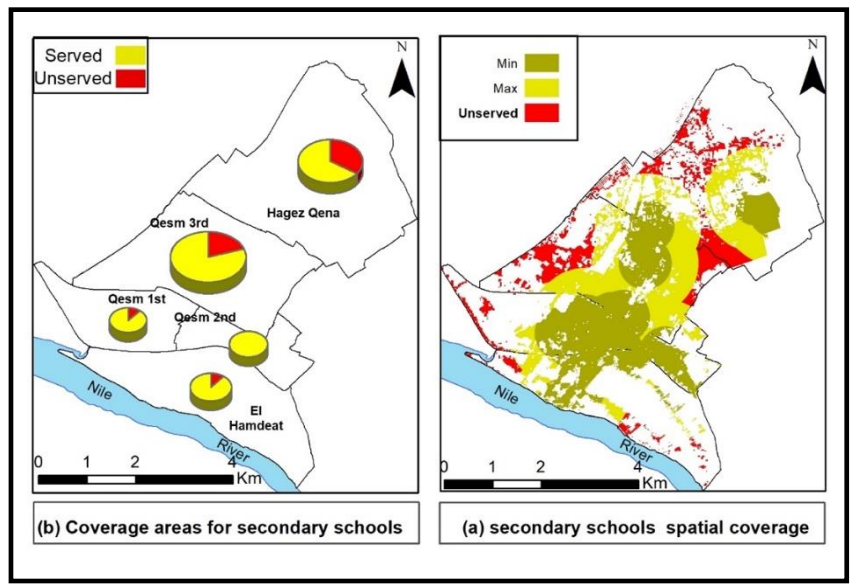

Figure (13): Assessment of secondary schools in Qena in 2021

Figure (13-a) shows the spatial coverage of secondary education schools in Qena from the built-up area according to the proposed standards. The serviced areas according to the standard represent 80% of the built-up area, while the unserved areas represent 20% of it. It includes South Valley University. Figure (13-b) shows the coverage areas for each administrative division. The Qesm2nd is the first with a coverage percentage of the serviced areas of 100%. Then comes (Al-Hemadat) and Qesm1st (87% each), followed by Qesm3rd (79%), and finally Hajar Qena (67%). Moreover, the unserved 0areas reached the maximum in (Hajar Qena) at 33%, and its minimum is in Qesm2nd.

*4.2.2. Health services*

Health services are among the necessary services concerned with the health of societies. The correlation between health and development indicators identifies that the health level matches the development level. It is essential as one of the critical social sectors that the state seeks to develop as an urban model. The government thus seeks to fulfil the highest quality of services for it and many population types at varying stations (Ghadban, 2013, p. 191). Hospitals are one of the essential basic services to preserve human life and increase their average lifespan (Halder et al.., 2020, p. 2581). Moreover, health care services are indicators of a civilized society and reflect its modernity. (Aziz M. A., Al-Helal H. I., 2020, p. 1.

- Ambulance Stations:

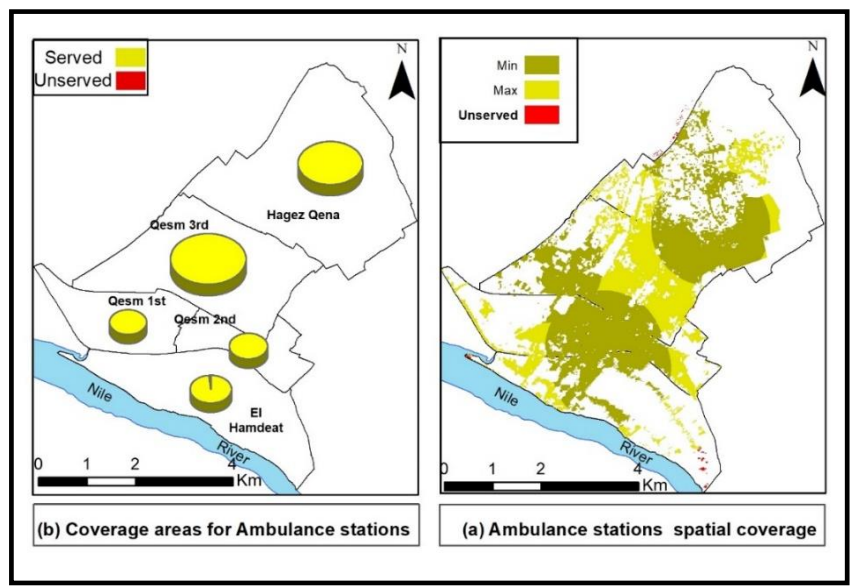

Figure (14): Assessment of ambulance service in Qena city in 2021

Figure (14-a) shows the spatial coverage of the ambulance stations in Qena from the built-up area according to the proposed standards. The serviced areas are about 99.8% of the built-up area, while the unserved areas occupy about: 0.2% in scattered spots (Al-Hamada). Figure (14-b) indicates the coverage areas for each administrative division. It indicates that the city's administrative divisions are fully serviced with ambulance stations, except for (Al-Hemadat), which has an unserved area representing 2% of the built-up area.

- Educational and Specialized Hospitals:

They represent educational and specialized hospitals with a coverage distance of 40-50 km, which makes the city fully served by health services and fills the deficit in the distribution of health units and centers.

- Health Units and Hospitals:

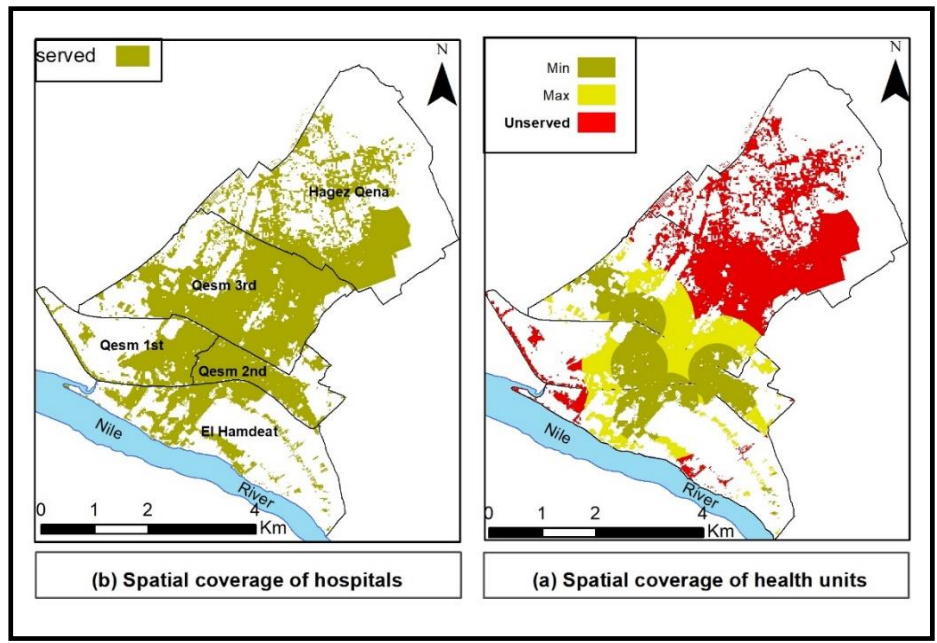

Figure (15): Assessment of health centers and hospitals in Qena city in 2021

Figure (15-a) shows the spatial coverage of health units in Qena city from the built-up area according to the proposed standards. The areas served according to the standard represent 50% of the total urban mass of the city. Figure (15-b) shows areas of spatial coverage for hospitals in Qena city from the built-up area. Furthermore, such areas appear to be fully serviced.

*4.2.3. Religious services*

- Mosques:

Figure (16-a) shows the spatial coverage of mosques in Qena from the built-up area according to the proposed standards. According to the standard, the serviced areas are about 86% of the built-up area, while the unserved areas are 14%, including South Valley University. Moreover, Figure (16-b) shows the coverage areas for each administrative division of the city of Qena. Qesm2nd is the first with a percentage of 99%, followed by Qesm1st and Qesm3rd with a percentage of 94%, followed by (Al-Hemadat) with a percentage of 89%, and finally (Hajer

Qena) with a percentage of 66% of the built-up area. The unserved areas reached the maximum in (Hajer Qena) with 34%, and the unserved areas reached the minimum in Qesm2$^{nd}$ at 1% of the built-up area.

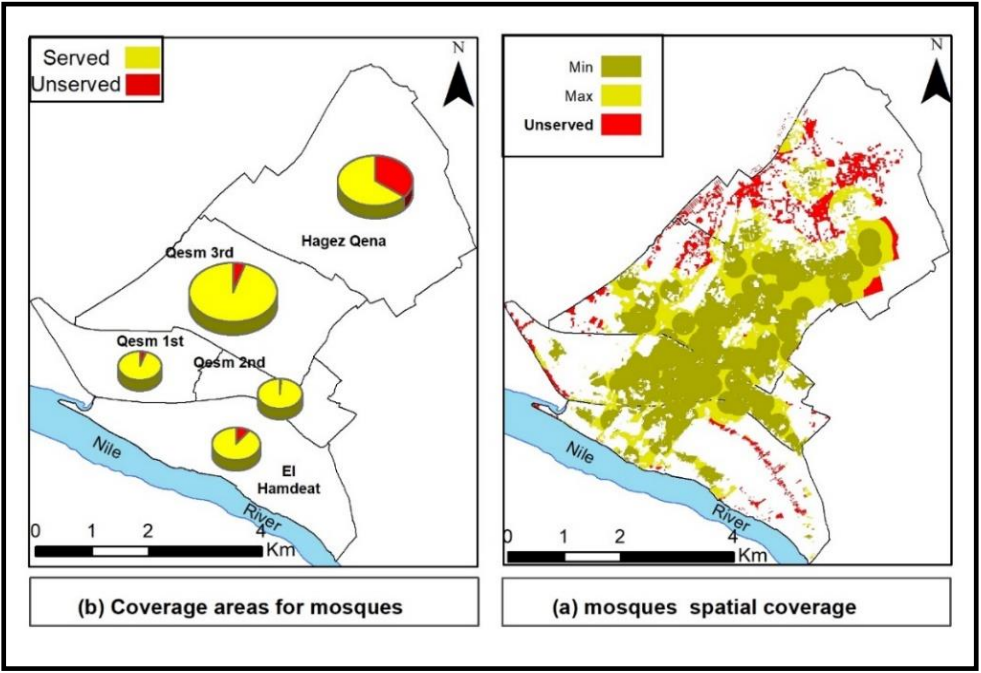

Figure (16): Assessment of mosques in Qena in 2021

- Churches:

Figure (17-a) shows the spatial coverage of churches in the city from the built-up area according to the proposed standards. The serviced areas according to the standard represent 78% of the built-up area, while the unserved areas represent 22% and include South Valley University. Figure (17-b) indicates the coverage areas for each administrative division. Accordingly, Qesm3$^{rd}$ is the first with a coverage percentage of the serviced areas at about 99%, followed by Qesm2$^{nd}$ with 95%, followed by (Al-Hemadat) with 93%, followed by Qesm1$^{st}$ with 89%, and finally (Hajar Qena) with 34% of the built-up area. Moreover, the unserved areas reached their maximum at (Hajer Qena) with 66% and their minimum in Qesm3$^{rd}$ with 1% of the built-up area.

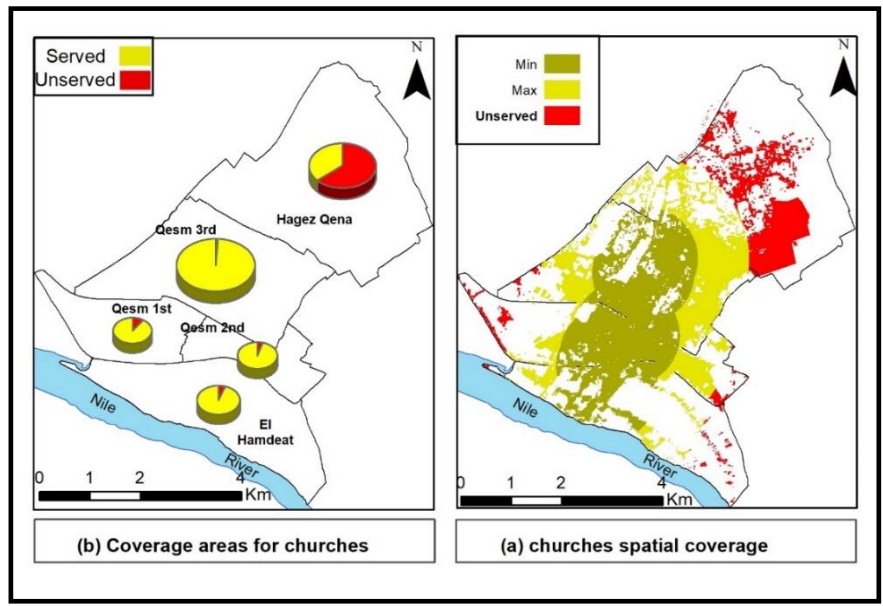

Figure (17): Assessment of churches in Qena in 2021

### 4.2.4. Cultural and Recreational services

cultural services refer to public libraries, cultural palaces, institutions, theatres, cinemas, museums, exhibitions, and local community institutions. Comprehensive planning for a city or region could not be done without considering the cultural identity of this community or region (Mashaqi, 2008, p. 37). Moreover, recreational services mean a set of activities and events that satisfy the desires of the population, and their psychological and mental comfort, according to their age and culture. Recreational services are among the critical and essential activities in the city and are necessary for all city plans (Ghadban, 2013, p. 228).

- Cultural Services:

Figure (18-a) shows the spatial coverage of cultural services in the city of Qena from the city's built-up area according to the proposed standards. The serviced areas according to the standard represent about 99.2% of the built-up area, while the unserved areas represent about 0.8% in the city's northern outskirts. Figure (18-b) shows the coverage areas for each administrative division, so all the administrative divisions of the city of Qena are provided with complete coverage of cultural services, except for (Hajar Qena), which has a coverage percentage of 97.4%.

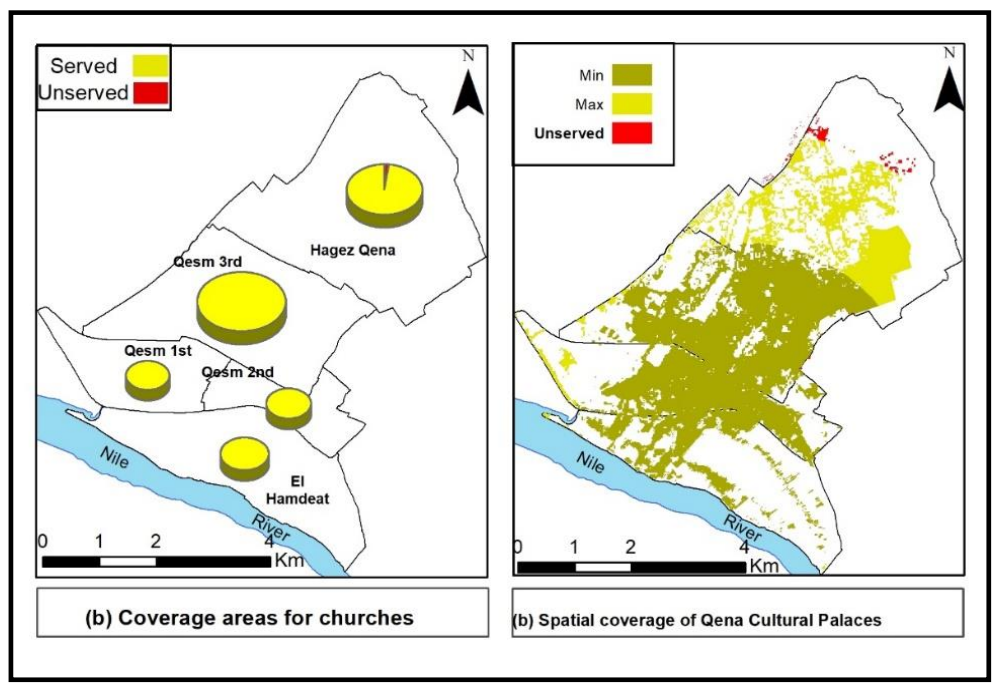

Figure (18): Assessment of cultural services in the city of Qena in 2021

- Recreational Services:

Table (2) the areas of parks and open spaces in the city of Qena

| Data | Al-Hemadeat (km$^2$) | Qesm1$^{st}$ (km$^2$) | Qesm2$^{nd}$ (km$^2$) | Qesm3$^{rd}$ (km$^2$) | Hajer-Qena (km$^2$) | Total Area (km$^2$) |
|---|---|---|---|---|---|---|
| Individual's share of parks and open areas | 0.520686 | 0.397804 | 0.368456 | 1.228874 | 0.102801 | 2.618624 |
| parks and open spaces | 0.153187 | 0.016685 | 0.005144 | 0.073085 | 0.007956 | 0.256057 |
| percentage of parks and open spaces (%) | 29.42 | 4.19 | 1.40 | 5.95 | 7.74 | 9.78 |
| Deficiency area | 0.367499 | 0.381119 | 0.363312 | 1.155789 | 0.094845 | 2.362567 |
| Deficiency Percentage (%) | 70.58 | 95.81 | 98.60 | 94.05 | 92.26 | 90.22 |

Source: based on the information infrastructure of the city of Qena - Population 2019, table (1).

Table (2) shows parks and open spaces and the unserved area of parks and open spaces in the city according to the proposed standards. According to the standard, the serviced areas represent about 10% of the total population share in the city, while the unserved areas represent about 90% of the city. Moreover, the administrative division of (Al-Hemadat) comes to the forefront to achieve about 29% of the total area required to meet the population's needs. In comparison, it comes in the lowest Qesm2$^{nd}$ with 1.4% of the total area required to meet the population's needs with the most significant deficit percentage of 98.6%.

*4.2.5. Other services*

- Postal services:

Figure (19-a) shows the spatial coverage of postal services in Qena from the built-up area according to the proposed standards. According to the standard, the served areas are about 91% of the built-up area. By contrast, the unserved areas reach about 9%, including scattered spots on the city's outskirts. Figure (19-b) indicates the coverage areas for each administrative division of the city of Qena. Accordingly, both Qesm2$^{nd}$ and Qesm3$^{rd}$ administrative divisions are the first with a coverage percentage of the serviced areas at 100%, followed by (Al-Hemadat) with a percentage of 95%, followed by Qesm1$^{st}$ with a percentage of 93%, and finally (Hajar Qena) at 75% of the built-up area. Furthermore, the unserved areas reached a maximum of 25% in (Hajer Qena), and the unserved areas reached their minimum in Qesm2$^{nd}$ and Qesm3$^{rd}$.

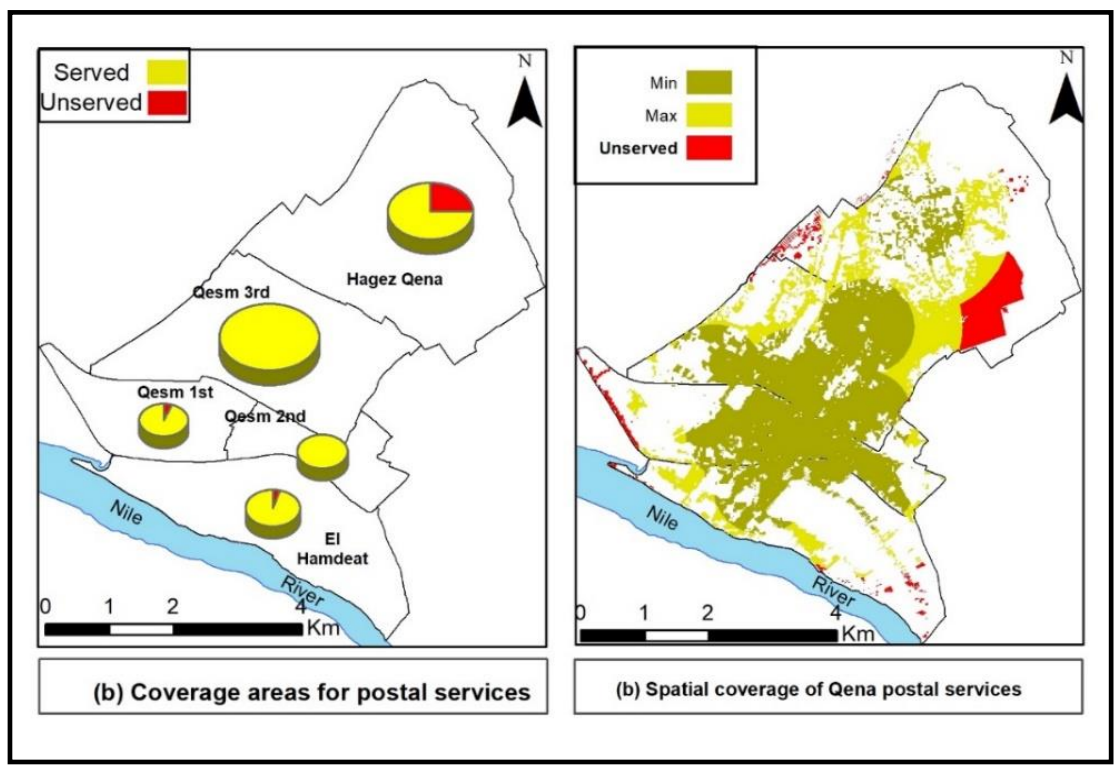

Figure (19): Assessment of the postal services in Qena city in 2021

- Fire Stations:

Figure (20-A) shows the spatial coverage of fire stations in Qena from the built-up area according to the proposed standards. The serviced areas according to the standard represent 99% of the built-up area, while the unserved areas represent 1%, represented by scattered places

on the outskirts of the city. Figure (20-b) indicates the coverage areas for each administrative division. Moreover, it indicates that the city's administrative divisions are fully serviced, except for (Al-Hemadat) with a percentage of 99%, which has an unserviced part representing 1% of the built-up area. In addition to Qesm1$^{st}$, which has a coverage percentage reaching 94% and has an unserved part representing 6% of the built-up area.

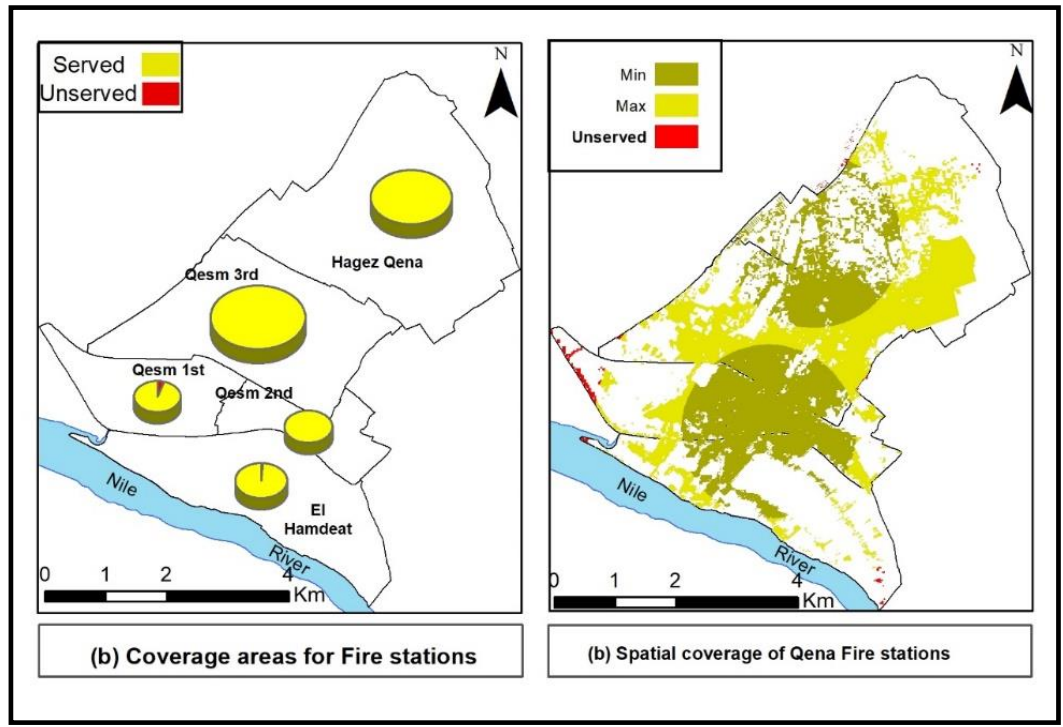

Figure (20): Assessment of fire stations in Qena city in 2021

### *4.3. Public Services General Indicators for intelligent spatial Analysis*

The importance of general indicators for evaluating services in the city of Qena is to give a comprehensive view of the distribution and coverage of services in the city. in addition, a comprehensive view reflects the distinguished areas of deficit and determines the percentage of each service.

#### *4.3.1. Services Heat Map*

Figure (21-a) shows the distribution of public services in the city of Qena and the extent of the concentration of services, as the Figure represents about 310 services distributed among the city's administrative divisions. Furthermore, the Heat map in Figure (21-b) shows the density of services, and their concentration in the midtown of the city of Qena, starting from 25 services/km$^2$ to reaching 45 services/km$^2$ and more. In comparison, it decreases towards the outskirts to reach less than five services/km², which shows the absence of spatial justice in the distribution of public services in the city. Thus, the remote areas of the city show poor localization of services.

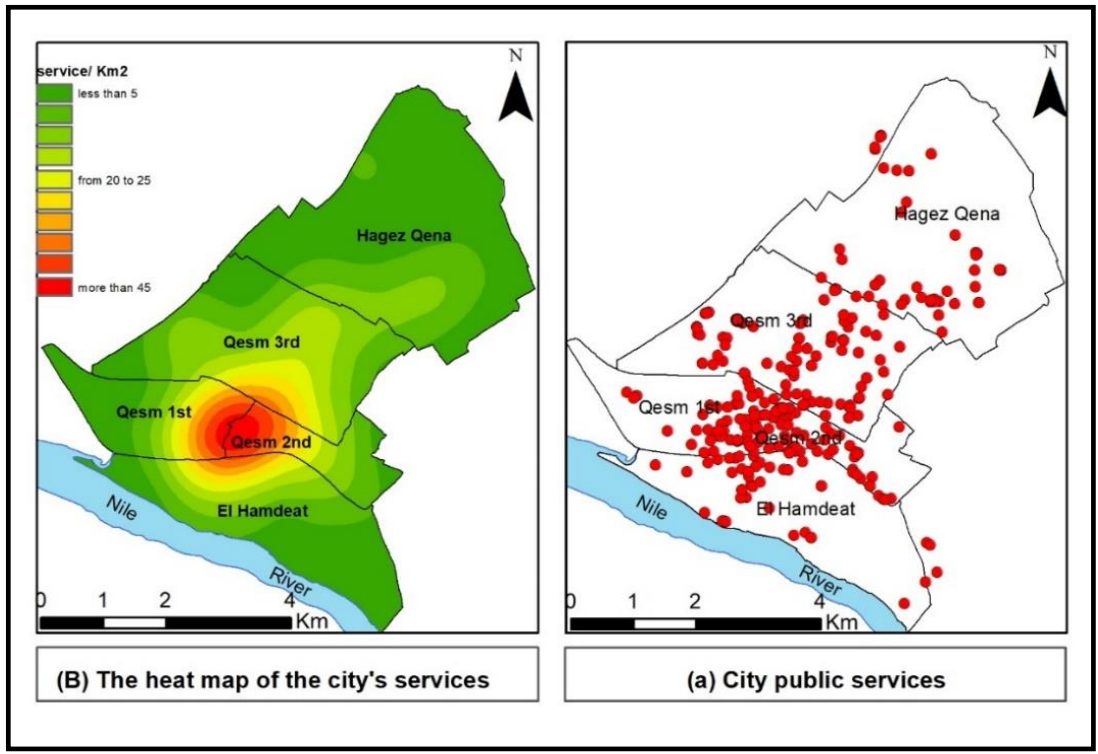

Figure (21): The heat map of services in Qena city in 2021

### 4.3.2. Services Coverage

Table (3) shows the coverage areas for public services in Qena and their percentages. The coverage percentage for services varies, reaching the maximum in hospitals. Moreover, the percentages of serviced areas vary until reaching the minimum in parks and open spaces, with about 9.78% of the built-up area. Additionally, the unserved areas vary, reaching the maximum in parks and open spaces with 90.22%, and the minimum in hospitals.

Table (3) served and unserved areas and their percentage in Qena City 2021

| service | Served area km² | % | Unserved area km² | % |
|---|---|---|---|---|
| kindergarten | 11.24 | 86.8 | 1.71 | 13.2 |
| primary | 11.29 | 87.2 | 1.66 | 12.8 |
| preparatory | 11.46 | 88.5 | 1.49 | 11.5 |
| secondary | 10.36 | 80 | 2.59 | 20 |
| Ambulance | 12.92 | 99.8 | 0.03 | 0.2 |
| Health units | 6.57 | 50.7 | 6.38 | 49.3 |
| Hospital | 12.95 | 100 | 0 | 0 |
| Mosques | 11.14 | 86 | 1.81 | 14 |
| Churches | 10.15 | 78.44 | 2.8 | 21.56 |
| Cultural | 12.85 | 99.2 | 0.1 | 0.8 |
| parks & open spaces | 256057 | 9.78 | 2,362,567 | 90.22 |
| postal | 11.83 | 91.4 | 1.12 | 8.6 |
| Fire stations | 12.84 | 99.15 | 0.11 | 0.85 |
| Average | 81.31% | | 18.69% | |

Source: based on the results of evaluating services according to the proposed standards, using PyCharm, ARC/GIS, Excel

Figure (22): Coverage percentage for services in Qena city in 2021

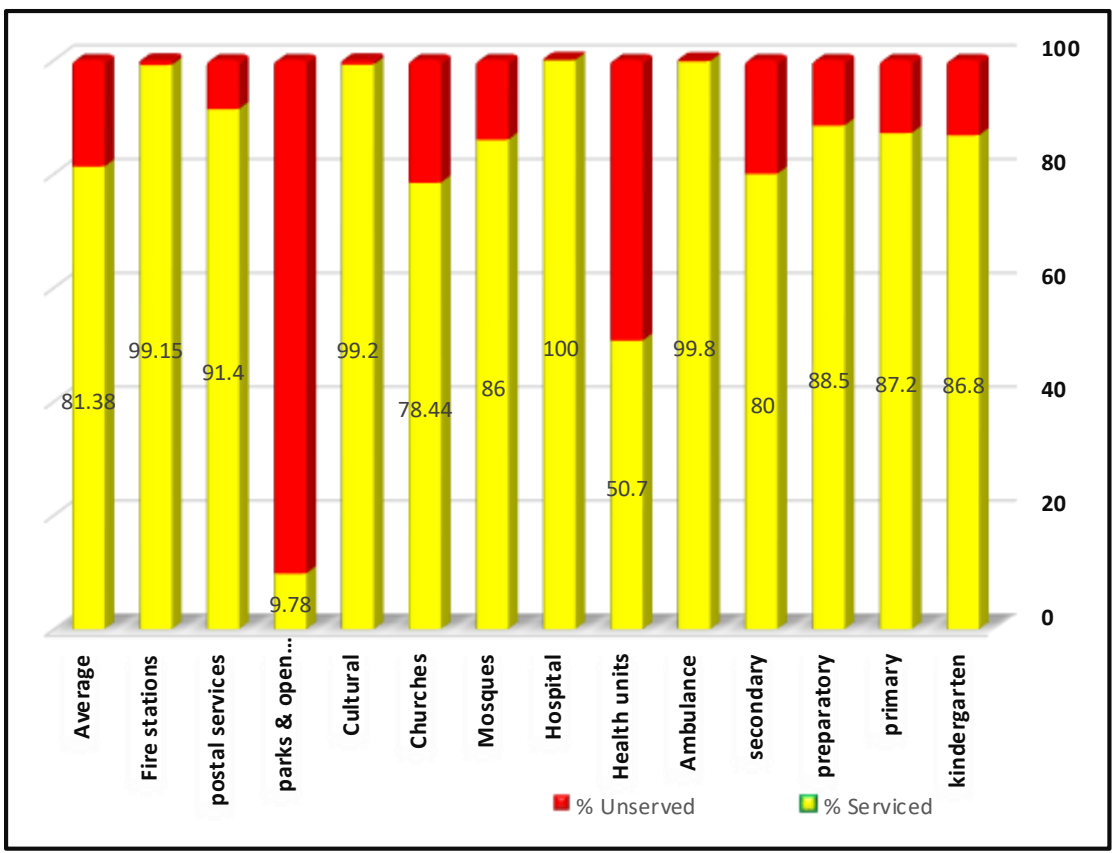

Figure (22) compares the coverage percentages for public services in Qena city and the variations between the coverage percentages for different services. Its spatial and statistical average can express the central tendency to cover services. The city's average spatial coverage of public services is about 81.3% and 18.7% for serviced and unserved areas, respectively. This shows that the general average of service coverage rates according to national standards is 82%, and the proposed standard for the city is 81.3%. For the health units, the city's general average is 50.7% for serviced areas due to the presence of hospitals that provide health services in the city along with health units. On the other hand, the percentage of parks and open spaces in the city differs from the general average of services. This is due to the severe deficiency in the average value of the individual's share in parks and open spaces.

## 5. Conclusion:

The study is concerned with providing intelligent spatial Analysis for public services in the city of Qena according to the catchment area of services to determine the most appropriate standards for each service. Then, it evaluates the services according to the proposed standards, using the intelligent spatial analysis algorithm using python, depending on both the descriptive and applied analytical approaches. In addition to building the algorithm based on the experimental approach. The effort to develop a suitable planning standard for the city was constrained by issues like how many factors should be taken into consideration and if the built-up area truly reflects the distribution of the population. The study has Indicated many results, the most important of which are shown below:

- Educational services: the areas serviced by kindergartens represent 87%, primary education 87%, preparatory education 88.5%, and secondary education 80% of the built-up area, respectively.
- Health services: the areas serviced by ambulance represent 99.8%, whereas the health units represent 50%, and hospitals show 100% of the built-up area.
- Religious services: The areas serviced by mosques represent 86%, and churches 78% of the built-up area respectively.
- Cultural and recreational services: The areas serviced by cultural services represent 99.2%, whereas parks and open spaces represent 10%.
- Other services: The areas serviced by postal services represent 91%, while firefighting points represent 99% of the built-up area.
- About two-thirds of the city's administrative divisions are compatible with the proposed standard, and public services are concentrated in the city's centre at a rate of more than 45 services/km$^2$. They decrease towards the outskirts to reach less than five services/km$^2$.
- The average coverage percentage for public services from the built-up area is about 81.3%, according to the local proposed planning standards for the city of Qena.

# 6. Sources and References:

## 6.1. Sources:

1. Information and Decision Support Centre in Qena Governorate, 2021, the information structure of Qena city, Qena: unpublished.
2. Central Agency for Public Mobilization and Statistics, 2017, General Population Census, Qena Governorate - Qena City.
3. General Organization for Urban Planning, 2016, Egyptian Planning Standards for Postal Services, Cairo.
4. General Organization for Urban Planning, 2014, Egyptian Planning Standards for Educational Services, Cairo.
5. General Organization for Urban Planning, 2016, Egyptian Planning Standards for Islamic Religious Services. Cairo.
6. General Organization for Urban Planning, 2015, Egyptian Planning Standards for Cultural Services, Cairo.
7. General Organization for Urban Planning, 2016, Egyptian Planning Standards for Christian Religious Services. Cairo.
8. General Authority for Urban Planning, 2014, Egyptian Planning Standards for Health Services, Cairo.
9. The Supreme Legislation Committee, 2019, Planning Standards for Public Services in the Emirate of Dubai, Dubai - United Arab Emirates: Government of Dubai - Official Gazette.
10. The General Administration of Urban Planning and Development, 2021, Digital Files of the city of Qena. Qena: unpublished.
11. Educational Buildings Authority, 2021, Digital Files of the city of Qena. Qena: unpublished.
12. Government of India, and Ministry of Urban Development-Town and Country Planning Organization. 2015. (URDPFI) GUIDELINES. Vol. 1. Mumbai, INDIA.

## 6.2. References:

1. Al-Dulaimi, Kh. H. A. (2015), Planning Community Services and Infrastructure Foundations and Standards (Second), Amman: Dar Al-Safaa.
2. Aziz. M. A. (2007), Applied Studies in Geographic Information Systems. Kuwait - Hawally: Dar Al-Ilm.
3. Aziz, M. A., & Al-Helal, H. I. (2020). Towards a Smart GIS Public Health Record System for the Capital Governorate, State of Kuwait. Journal of Community Medicine & Public Health, 4(03), 1–16. https://doi.org/10.29011/2577-2228.100094
4. Erickson, J. (2019). Algorithms-UIUC. Jeff Erickson.
5. General Authority for Urban Planning. (2014). Egyptian planning standards for educational services. Cairo: General Authority for Urban Planning.
6. Ghadban, F. B. (2013), The Geography of Services. Amman: Dar Al-Yazuri.
7. Mashaqi, A. (2008), Analysis and Evaluation of the Distribution of Health, Education, Cultural and Recreational Services in Nablus Governorate. Nablus, Palestine: An-Najah National University, an unpublished Master's thesis.
8. Goodchild, M. F. (2017). Big Geodata. Comprehensive Geographic Information Systems, 3, 19–25. https://doi.org/10.1016/B978-0-12-409548-9.09595-6

9- Halder, B., Bandyopadhyay, J., & Banik, P. (2020). Assessment of hospital sites' suitability by spatial information technologies using AHP and GIS-based multi-standards approach of Rajpur–Sonarpur Municipality. Modeling Earth Systems and Environment, 6(4), 2581–2596. https://doi.org/10.1007/s40808-020-00852-4
10- Kennedy, H. (2000). The ESRI Press Dictionary of GIS Terminology. Redlands-California: ESRI Press.
11- Nuzir, F. A., & Dewancker, B. J. (2014). Understanding the Role of Education Facilities in Sustainable Urban Development: A Case Study of KSRP, Kitakyushu, Japan. Procedia Environmental Sciences, 20, 632–641. https://doi.org/10.1016/j.proenv.2014.03.076
12- Richard Szeliski. (2020). Texts in Computer Science. In Algorithms and applications (Third, Vol. 42). Cham, Switzerland: Springer Nature Switzerland.
13- Tateosian, L. (2015). Python for ArcGIS. In Python for ArcGIS. London, UK: Springer International Publishing. https://doi.org/10.1007/978-3-319-18398-5

### 6.3. Websites:

1- ArcGIS. Retrieved May 22, 2021, from www.arcgis.com
2- ArcPy. Retrieved May 29, 2021, from https://desktop.arcgis.com/en/arcmap/10.7/analyze/arcpy/what-is-arcpy-.htm#
3- ESRI. Retrieved May 6, 2021, from https://www.esri.com/en-us/home
4- PyCharm. Retrieved March 5, 2021, from https://www.jetbrains.com/pycharm/
5- Python. Retrieved May 23, 2021, from https://www.python.org/
6- Python packages. Retrieved May 27, 2021, from https://pypi.org/
7- Stack overflow. Retrieved May 1, 2021, from https://stackoverflow.com/